\documentclass{article}

     \usepackage[preprint]{neurips_2022}

\pdfoutput=1
\usepackage[utf8]{inputenc} 
\usepackage[T1]{fontenc}    
\usepackage{hyperref}       
\usepackage{url}            
\usepackage{booktabs}       
\usepackage{amsfonts}       
\usepackage{nicefrac}       
\usepackage{microtype}      
\usepackage{xcolor}         

\usepackage[pdftex]{graphicx}
\usepackage{subfig}
\usepackage{multirow}
\usepackage{mathtools}
\usepackage{wrapfig}

\usepackage[noend]{algpseudocode}
\usepackage{float}

\title{Transferable Physical Attack against Object Detection with 
	Separable Attention}

\author{%
  Yu Zhang$^{1}$, Zhiqiang Gong$^{2}$, Yichuang Zhang$^{1}$, YongQian 
  Li$^{1}$, \\
  \textbf{Kangcheng Bin$^{1}$, Jiahao Qi$^{1}$, Wei Xue$^{3}$, Ping 
  	Zhong$^{1}$}
  \thanks{Ping Zhong is the corresponding author. }
 \\
$^{1}$National University of Defense Technology, $^{2}$Defense Innovation 
Institute,\\ 
Chinese Academy of Military Science, $^{3}$Anhui University of 
Technology\\
\texttt{\{zhangyu13a, gongzhiqiang13, ycz, liyongqian, binkc21,  }\\ \texttt{
qijiahao19, 
zhongping\}@nudt.edu.cn}, 
xuewei@ahut.edu.cn \\
}

\begin{document}

\maketitle

\begin{abstract}
  Transferable adversarial attack is always in the spotlight since deep 
  learning models have been demonstrated to be vulnerable to adversarial 
  samples. 
  However, existing physical attack methods do not pay enough attention on 
  transferability to unseen models, thus leading to the poor 
  performance of black-box attack.
  In this paper, we put 
  forward a novel method of generating  physically realizable adversarial 
  camouflage to achieve transferable attack against detection models. More 
  specifically, we first introduce multi-scale attention maps based on 
  detection models  to capture features of 
  objects with various resolutions. 
  Meanwhile, we adopt a sequence of composite transformations to obtain the 
  averaged attention maps, which could curb model-specific  noise in the  
  attention
  and thus further boost transferability. 
  Unlike the general visualization interpretation methods where model 
  attention should be put on the foreground object as much as possible, 
  we carry out attack on separable attention from the 
  opposite perspective, 
  \textit{i.e.} suppressing attention of the foreground and enhancing that of 
  the background. Consequently, transferable adversarial camouflage could be 
  yielded efficiently with our novel attention-based loss function. Extensive 
  comparison experiments 
  verify the superiority of our method to state-of-the-art 
  methods. 
\end{abstract}

\section{Introduction}

In recent years, deep neural networks (DNNs) have achieved a great 
success among various scenarios, \emph{e.g.} object detection 
\cite{redmon2018yolov3,zhu2020deformable,ren2015faster}, image 
classification \cite{gong2019cnn,gong2020statistical} and image 
segmentation \cite{he2017mask}. However, it has been found that DNNs 
are vulnerable to adversarial samples \cite{goodfellow2014explaining} and the 
predicted results may be 
distorted dramatically by crafting  human-imperceptible perturbations 
towards the input samples.

Plenty of methods have been proposed about how to generate adversarial 
samples \cite{wu2020boosting,wang2021feature,huang2020universal}, which could 
be classified into \emph{white-box} vs. \emph{black-box} 
attack according to the transparency of model information. \textit{White-box} 
attack permits intruders to get access 
to model structure, model parameters and even the training dataset when 
generating adversarial samples. However, this information would be 
inaccessible for 
\textit{black-box} attack.
Actually, adversarial attack could also 
be categorized as \emph{untargeted} vs. \emph{targeted} 
attack and \emph{digital} vs. \emph{physical} 
attack.  \textit{Untargeted} attack is conducted with the 
aim of altering model prediction only, not caring about what the final 
predicted label is.
On the contrary, 
\textit{targeted} attack requires that the prediction  should be misclassified 
to a specific label. \textit{Digital} attack assumes that the adversary can 
manipulate image pixels in the digital space, where perturbation constrains 
(\emph{e.g.} $L_{0}$ norm \cite{su2019one} or $L_{2}$ 
norm \cite{yang2020learning}) would be adopted to avoid being suspected.
As for \textit{physical} attack, it 
tries to modify objects and confuse DNNs in the physical world or relevant
simulation environment, where a host of practical influence factors 
should be taken into account, making it much more difficult and challenging 
than that of digital attack \cite{wang2021dual,jiang2021fca}. For the 
sake of practicality, this paper mainly focuses on untargeted physical 
black-box attack.  

\begin{figure}
	\centering
	\includegraphics[scale=0.185]{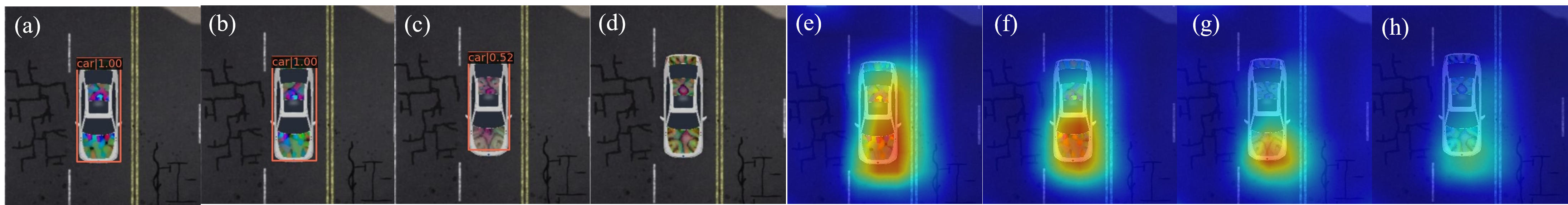}
	\caption{Typical detection results and attention maps of vehicle covered 
		with different 
		adversarial patterns. All the patterns are generated on Yolo-V3 and 
		tested 
		on Faster R-CNN. (a)$\sim$(c) are patterns generated by adopting 
		adversarial 
		losses from FCA \cite{jiang2021fca}, DAS \cite{wang2021dual} and 
		ATA \cite{wu2020boosting} respectively (refer to subsection 
		\ref{Comparison} for 
		more details). (d) displays the 
		camouflage manufactured with our method. (e)$\sim$(h) are 
		corresponding attention 
		maps 
		of (a)$\sim$(d). 
	}  
	\label{tranferability}
\end{figure}

Although a great quantity of attack methods have sprung up, the limitations 
still exist: 1) The majority of methods are about
classification 
attack \cite{wu2020boosting,wang2021feature,chen2020universal},
which could not be applied to object detection directly.
2) It is 
difficult for existing 2D adversarial  physical 
patch \cite{eykholt2018robust,thys2019fooling} to adapt to multi-view scenario 
of 3D space. 
3) Limited attention was paid on transferability for previous physical attack 
methods 
\cite{maesumi2021learning,huang2020universal,wang2021dual}, 
which would lead to the degeneration of black-box attack. 
To address those issues,
our work focuses on crafting a physically realizable 3D adversarial pattern 
that could  deceive black-box detectors with  high transferability.
Specifically, the pattern is initialized given the sampled faces of a 3D 
vehicle model mesh, and then optimized by attacking the averaged multi-scale 
attention maps.
Besides, unlike the general middle-layer attack 
strategies \cite{wu2020boosting,wang2021dual}  which 
take the map as a 
whole during optimization, we carry out attention attack by considering the 
contribution of foreground and background attention respectively. 
Consequently,  a physical structured camouflage with high transferability  
could be yielded with  our  proposed method.
Typical  comparisons  about detection results and attention distraction are 
displayed in Figure \ref{tranferability}, which shows that our method has 
better performance than baseline methods, \textit{i.e.} our adversarial 
texture could bring about the greatest reduction in  both detected class 
probability  and model attention.

Our contributions include the following:1) To the best of our knowledge, we 
are the first to conduct
Attack on Separable Attention (ASA) with a novel attention-based loss 
	function, which is mainly made up of two  modules, Foreground 
	Attention Suppression module and Background Attention Amplification 
	module.	
	2) Based on detection models, multi-scale attention maps are developed to 
	capture features of objects at different resolutions. In order to further 
	boost transferability, the maps are 
	smoothed with a sequence of compound transformations, which could 
	alleviate model-specific noise. 	
	3) Experiments show that our method has a good performance and  
	outperforms other state-of-the-art 
	approaches in terms of black-box transferability.

\section{Related work}
\label{Related_work}
\paragraph{Physical Adversarial Attacks}
The purpose of physical adversarial 
attack is to craft localized visible perturbations that have the potential to 
deceive DNN-based vision systems. According to space 
dimension, physical attack could be classified into 2D physical 
attack 
\cite{eykholt2018robust,thys2019fooling,sharif2016accessorize,wu2020making,wang2021towards,du2022physical}
 and 3D physical 
attack 
\cite{athalye2018synthesizing,jiang2021fca,maesumi2021learning,wang2021dual}.
Sharif \textit{et al.} \cite{sharif2016accessorize} developed a method of 
deceiving face-recognition system by generating physical eyeglass frames. 
Eykholt \textit{et al.} \cite{eykholt2018robust} came up with Robust 
Physical Perturbations (RP2), a general attack method that could 
mislead  classification of stop sign under different environmental conditions. 
Thys \textit{et al.} \cite{thys2019fooling} paid attention to 
physical attack against person in the real world by putting one 2D adversarial 
patch on torso. The method presented a good performance in the front view, but 
cannot work well for larger shooting 
angles \cite{tarchoun2021adversarial}.  
Fortunately, this problem could be overcome by 3D physical attack.
Athalye  \textit{et al.} \cite{athalye2018synthesizing} developed Expectation 
Over Transformation, the first approach of generating 3D robust 
adversarial samples. Maesumi \textit{et al.} \cite{maesumi2021learning} 
presented a universal 3D-to-2D adversarial attack method, where a structured 
patch was sampled from the reference human model and the human pose could be 
adjusted freely during training.
Wang \textit{et al.} \cite{wang2021dual}  proposed Dual Attention 
Suppression (DAS) attack based on an open source 3D virtual environment, 
and Jiang  \textit{et 
	al.} \cite{jiang2021fca} extended DAS by presenting a full-coverage 
	adversarial 
attack.

\paragraph{Black-box Attack}

The intent of black-box attack is to yield a pattern that not only has 
a striking effect in white-box attack, but can misguide black-box models with 
high transferability. Generally, black-box attack could be divided into 
query-based attack and transfer-based attack.  The former method 
estimates gradient roughly by obtaining the variation of model output with 
disturbed inputs, and then updates adversarial samples like white-box 
attacks \cite{yang2020learning}\cite{xiang2021local}\cite{guo2019simple}. Due 
to the complexity of DNNs, plenty of queries are normally inevitable (which 
might be 
impermissible in the real world) if one 
attacker wants to get a better gradient estimation. The latter method, 
\textit{i.e.} 
transfer-based attack, relies on transferability of adversarial 
sample 
\cite{zhou2018transferable,inkawhich2019feature,wang2021feature,huang2019enhancing,inkawhich2020perturbing}.
Zhou \textit{et al.} \cite{zhou2018transferable} displayed that 
attack performance could be improved by enlarging the distance of intermediate 
feature maps between benign images and the corresponding adversarial samples. 
Inkawhich \textit{et 
	al.} \cite{inkawhich2019feature} found that one particular layer may play 
a more important role than others in adversarial attack. 
Feature importance-aware attack was proposed in \cite{wang2021feature}  with 
the purpose of enhancing model transferability by destroying crucial 
object-aware features. 
Chen \textit{et 
	al.} \cite{chen2022relevance} realized transferable 
attack on object detection by manipulating relevance maps. Specifically, 
they first computed two attention maps corresponding to the original class and 
another class respectively, and then minimized the ratio of them. Although 
their 
method works for multi-class datasets, it is difficult to generate heatmaps of 
the second class if there is only one category to be detected. Wang \textit{et 
	al.} \cite{wang2021dual} conducted attack by
shattering the "heated" region with a recursive 
method. However, our experiments show 
that it would be inefficient to search for the "heated" regions for some 
high-resolution maps.

\section{Proposed Framework}
\label{Proposed_Framework}

\begin{figure*}
	\centering
	\includegraphics[scale=0.32]{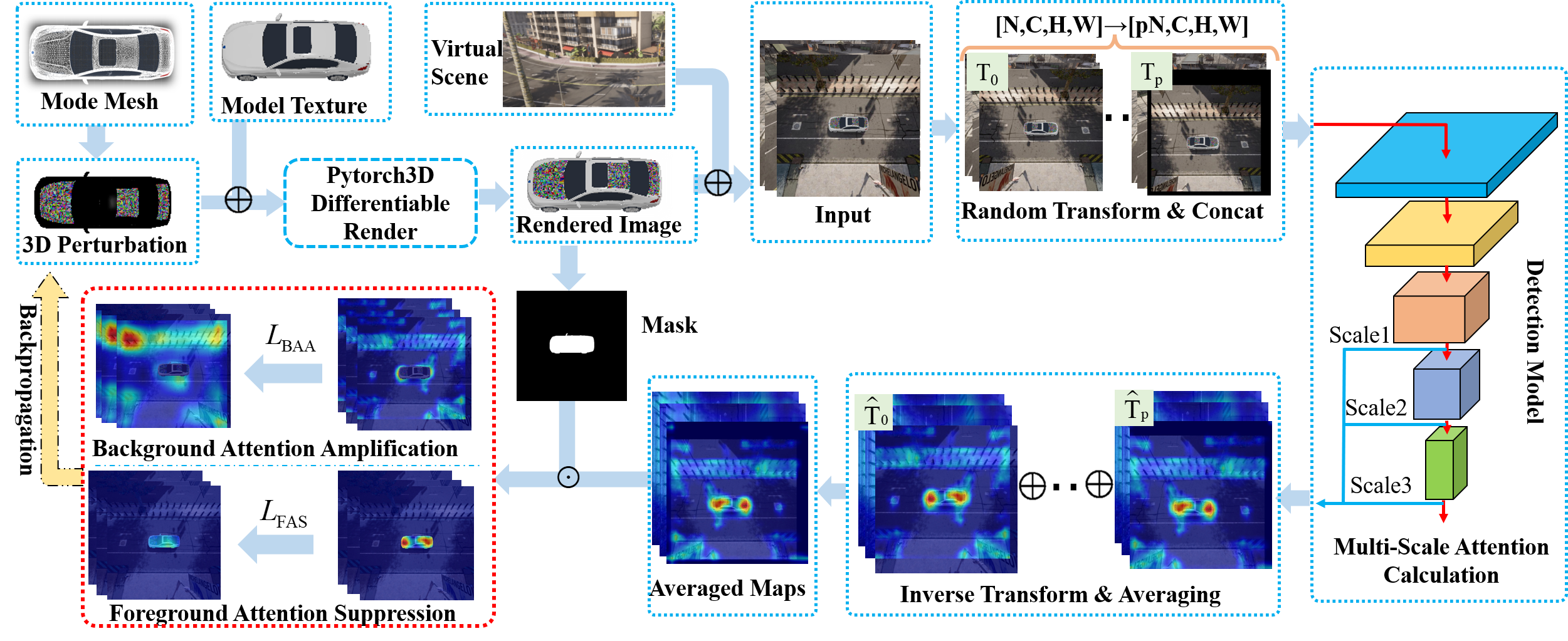}
	\caption{The proposed pipeline. We first sample a set of triangular 
		faces from mesh, then initialize them with random noise, and finally 
		render the 3D model on virtual background. After that, several 
		compound 
		transformations are carried out towards the original inputs before 
		fed  
		into detection model, where multi-scale attention maps 
		can be obtained. 
		To suppress model-specific noise of the maps, we align the maps with 
		original 
		input images by several inverse transformations, and subsequently 
		obtain
		averaged map by linearly combining the aligned maps. Finally, the 
		perturbation 
		pattern is optimized by suppressing foreground attention and 
		amplifying background attention. 	  
	}\label{mypipeline}

\end{figure*}

In this section, we present a framework of  physically realizable  adversarial 
attack on object detection. The main purpose of our work is to generate a 
structured 3D 
adversarial pattern such that, when pasted 
on the surface of one vehicle, the pattern 
can fool black-box detectors with high probability. 
The overall pipeline of our 
method is shown in Figure \ref{mypipeline}.

\subsection{Preliminaries}

Unlike 2D adversarial attack, the input image in 3D physical attack is 
ususally a 
rendered result of one object model based on renderer $\mathcal{R}$, 
\emph{i.e.} $\mathcal{R}\left [ \left ( M,T \right ),P \right ]$,
where $M$ denotes object mesh that consists of a large number of triangular 
faces, $T$ is the corresponding texture map and $P$ represents camera 
parameter.
This paper adopts a modular differentiable 
renderer provided by Pytorch3D \cite{ravi2020accelerating}.
Since the
rendered 2D image is made up of foreground object and a totally white 
background, we can get the object mask $m$ easily by setting  values of 
background pixels to zeros and the remaining pixels to ones. In order to 
obtain a physical world background $B$, we take CARLA 
\cite{dosovitskiy2017carla} as our 3D virtual simulation environment just like 
previous work of Wang \textit{et al.} \cite{wang2021dual}. Then the synthetic 
2D input $X$ based on 3D 
model could be obtained as below:
\begin{equation}
	X=m\odot   \mathcal{R}\left [ \left ( M,T \right ),P \right ] +   
	(1-m)\odot   B \label{1}
\end{equation}

For one detection model $f_{\theta }\left ( \cdot  \right )$  parameterized by 
$\theta$, the aim of physical attack is to update pattern $T$ that can mislead 
the model by optimizing the following equation:
\begin{equation}
	T_{adv} =\underset{T}{\mathrm{arg}\,  \mathrm{min}} \,  \mathcal{L} \left 
	( 
	f_{\theta }\left ( X  \right ),G   \right ) \label{2}
\end{equation}
where $G$ indicates object ground truth and $\mathcal{L}$ means a specific 
loss function that encourages the object detector to generate an incorrect 
output with either a misaligned bounding box or a low class (object) 
confidence. To realize high transferable attack, our work develops a novel  
attention-based loss function, which will be presented in the following 
subsections.

\subsection{Generating  Multi-scale Attention Maps with Detection 
		Model}

Motivated by Grad-CAM \cite{selvaraju2017grad}, we propose a
detection-suitable and noise-reduced  attention generation method to make it 
qualified to transferable attack on object detection.
As regards how to get attention maps, the original Grad-CAM first calculates 
the gradients of $y_{c}$ (the predicted probability of class $c$) with 
respect to feature maps of certain activation layer $A^{l}(X)$, and then 
accumulate 
feature maps with gradient-based weights. However, unlike classification 
models which only output a vector of 
probabilities corresponding to different classes, object detection models 
usually yield a certain number of vectors including the predicted bounding 
boxes and probabilities. 

Taking Yolo-V3 \cite{redmon2018yolov3} as an example, 
which is also one of the white-box models in our experiment, the model 
generates detection result of
$\mathbf{O} =\left ( \mathbf{b_{x}}, \mathbf{b_{y}}, 
\mathbf{b_{w}},\mathbf{b_{h}},\mathbf{C_{obj},\mathbf{P_{cls}}} \right ) $  
before Non-Maximum Suppression (NMS), 
where $\mathbf{b_{x}}$,$\mathbf{b_{y}}$,$\mathbf{b_{w}}$ and $\mathbf{b_{h}}$ 
are parameters of the predicted boxes,
 $\mathbf{C_{obj}}$ indicates object confidence 
and $\mathbf{P_{cls}}$ represents class probability. It should be noticed that 
every element of $\mathbf{O}$ is a concatenated vector of  values  
corresponding to all the candidate objects. Similar to the original Grad-CAM, 
we have to first 
figure out what content of $y_{c}$ includes.  For one-stage detection model 
with 
input $X$,  $y_{c}$ is chosen as follows: 
\begin{equation}
	y_{c}\left ( X \right ) =max\left ( \mathbf{C_{obj}}+\mathbf{P_{cls,c}} 
	\right 
	)  
	\label{3}
\end{equation}
where $\mathbf{P_{cls,c}}$ denotes the probability vector of class $c$. It is 
worth noting that we do not consider NMS because it may yield null values 
with adversarial samples. With regard to two-stages detectors like Faster 
R-CNN, we just choose 	
$y_{c}\left ( 
X \right ) =max\left (\mathbf{P_{cls,c}+IOU_{b,B}} \right )$, where 
$\mathbf{IOU_{b,B}}$ denotes intersection over union between detected boxes 
and ground truth boxes. 
Then we can get a coarse map of layer $l$ 
by linearly combining activation maps and corresponding weights  in 
channel direction:
\begin{equation}
s^{l}=\mathrm{ReLU}\left (\frac{1}{Z} \cdot\sum_{k} \sum_{i} \sum_{j} 
\frac{\partial y_{c}(X)}{\partial A_{i, j}^{k, l}(X)}\cdot A^{k,l}  \right )
	\label{4}
\end{equation}
where  $k$ represents channel index, $(i,j)$ denotes pixel index of 
activation map 
$A^{k,l}$ and $Z$ is the 
total number of pixels in $A^{k,l}$ . 
For simplicity purposes, the symbol of class $c$ is 
ignored in the calculated result. It is obvious that 
the dimension of attention map in equation (\ref{4}) is equal to that of   
$A^{k,l}$, which would vary with the location of layer.  To merge the  
attention of different layers, the maps should be adjusted  to align with 
input images.

In general, the produced attention maps based on the above procedure would 
bring 
about model-specific noise, which may hinder the transferability of attack 
among different models. To alleviate this issue, we first introduce several 
compound 
transformations $T_{i}$ to the input $X$ of formula (\ref{1}):
\begin{equation}
	\widetilde{X}_{i}=T_{i}(X)=t_{i,q}\left(\cdots t_{i,0}(X)\right), \quad 
	i=0, 
	\cdots, 
	p \label{5}
\end{equation} 
where $p$ is the total number of compound transformations, and transformation 
$T_{i}$ consists of a total number of $q$ base random transformations 
$t_{i,0} \cdots t_{i,q}$  (including horizontal flip, translation, scaling, 
\textit{etc.}). After 
that, the transformed inputs $\widetilde{X}_{i}$  are concatenated in batch 
dimension before fed into detectors. Subsequently, we can get the 
corresponding attention map $s_{i}^{l}$ according to formula 
(\ref{3})$\sim$(\ref{4}).
As the generated attention $s_{i}^{l}$ would be transformed simultaneously due 
to 
the adopted random transformations, so we need to introduce 
inverse transformations to make  $s_{i}^{l}$ aligned with each 
other:  
\begin{equation}
	\widetilde{s}_{i}^{l}=\widehat{T}_{i}\left(s_{i}^{l}\right)=\widehat{t}_{i,0}\left(\cdots
	\widehat{t}_{i,q}\left(s_{i}^{l}\right)\right), \quad i=0, \cdots, p
	\label{6}
\end{equation} 
where $\widetilde{s}_{i}^{l}$ denotes the aligned attention map and 
$\widehat{T}_{i}$ is compound inverse transformation that consists of 
serial inverse base transformations $\widehat{t}_{i,0} \cdots 
\widehat{t}_{i,q}$ (corresponding to $t_{i,0} \cdots t_{i,q}$ respectively).

In order to merge different attention maps, we propose an efficient and direct 
method as below:     
\begin{equation}
	S^{l}=\frac{1}{p} \sum_{i=0}^{p}\widetilde{s}_{i}^{l} , \quad 
	l=1,\cdots,N  
	\label{7}
\end{equation} 
where $S^{l}$ is the averaged attention map of layer $l$ and $N$ is the total 
number of 
attention scales. One advantage of equation (\ref{7}) is that the 
noise could be smoothed without extra forward propagation, which can enhance 
the efficiency of optimization procedure remarkably.

Following the processes above, the averaged attention map could be 
yielded for any convolutional layer. 
Intuitively, attention of a single layer 
may not be enough to capture objects at different scales. 
Especially, the attention of a certain layer may even vanish for some small 
objects due to the convolution operation in the encoding process.
Thus, it is necessary to combine 
multi-scale attentions to capture features of objects with
various resolutions.

\subsection{Attacking on Separable Attention (ASA)}
Unlike previous attention-based attacks which take the generated map 
as a whole, we propose to attack on separable attention, \emph{i.e.} our 
method 
suppresses attention of the foreground and amplifies that of the background 
simultaneously. Therefore, our novel attention 
loss $\mathcal{L}_{A}$  is made up of two parts and can be formulated as 
follows:
\begin{equation}
	\mathcal{L}_{A}= \mathcal{L}_{FAS}+\mathcal{L}_{BAA}
	\label{8}
\end{equation}
where $\mathcal{L}_{FAS}$  denotes Foreground Attention Suppression ($FAS$) 
loss and $\mathcal{L}_{BAA}$ indicates Background Attention Amplification 
($BAA$) loss.
Apparently, those losses are closely related to  the foreground attention 
$S^{f}$ and the background attention $S^{b}$, which could be obtained with the 
object mask $m$:
\begin{equation}
	S^{f}= \sum_{l=1}^{N} S^{l}\odot m  
	,\quad  
	S^{b}= \sum_{l=1}^{N} S^{l}\odot 
	\left ( \mathbb{I} -m \right )  
	\label{9}
\end{equation}
where $S^{l}$ denotes attention map of layer $l$ deduced in the 
last subsection, $\mathbb{I}$ indicates some all-1 matrix with the same
dimension as mask $m$, and the summation on the right side hints that the 
separable maps cover different scales of attention.

To alter the distribution of foreground and background attention, one direct 
and efficient method is to change their global average values, which are 
defined as $\overline{S^{f}}$ and $\overline{S^{b}}$ respectively: 
\begin{equation}
	\overline{S^{f}}  =\frac{\sum_{i,j}S_{i,j}^{f}}{\left \| m \right \|_{0}  }
	,\quad
		\overline{S^{b}}  =\frac{\sum_{i,j}S_{i,j}^{b}}{\left \| \mathbb{I}-m 
		\right 
		\|_{0}  } 	
	\label{10}
\end{equation}
where $S_{i,j}^{f}$ and $S_{i,j}^{b}$ are pixel values of foreground and 
background attention at position $(i,j)$, and $\left \| \cdot  \right \| _{0} 
$ means the total number of nonzero elements. 
Actually, adversarial texture can be optimized by minimizing  
$\overline{S^{f}}$ and  maximizing  $\overline{S^{b}}$ simultaneously.  

However, a low (high) global average value of pixels does not guarantee that 
all the local pixels have low (high) values. To solve the problem, we 
define $\varphi \left ( \overline{S^{f,b}},k\right )$ by 
sorting all the pixels of  $\overline{S^{f}}$ or  $\overline{S^{b}}$ in 
descending order, taking out the top $k$ values and calculating their 
mean value.
Moreover, the ratio of $\varphi 
\left ( \overline{S^{f,b}},k\right )$ to $\overline{S^{f,b}}$ is adopted 
to avoid the situation where attention loss becomes too small to 
guide the optimization. By doing so, 
the ratio could be always larger than 1 unless all pixels have the same 
value. Thus, we can get $\mathcal{L}_{FAS}$ and $\mathcal{L}_{BAA}$ as 
below:  
\begin{equation}
	\mathcal{L}_{F A S}=\alpha_{1} \cdot \overline{S^{f}}+\alpha_{2} \cdot 
	\frac{\varphi\left(\overline{S^{f}}, k\right)}{\overline{S^{f}}}
	,\quad
		\mathcal{L}_{B A A}=-\alpha_{1} \cdot \overline{S^{b}}-\alpha_{2} 
		\cdot 
	\frac{\varphi\left(\overline{S^{b}}, k\right)}{\overline{S^{b}}}
	\label{11}
\end{equation}
where $\alpha_{1}$ and $\alpha_{2}$ are 
hyperparameters, and the minus sign in $\mathcal{L}_{B A A}$ implies that we 
expect the background attention to be enhanced during optimization.

\subsection{The Overall Algorithm}

To promote the naturalness of the 
pattern, we resort to 3D smooth loss \cite{maesumi2021learning} showed 
as below:
\begin{equation}
	\mathcal{L}_{S,3D}=\sum_{e\in E }\left | e \right |\cdot \left | C\left ( 
	F_{1}  \right ) - C\left ( F_{2}  \right ) \right |  
	\label{12}
\end{equation}
where  $C\left ( F \right )$ is  RGB color vector of $F$, $E$ represents all 
the edges of sampled triangular faces except for boundary edges, $| e |$ is 
the length of edge $e$, $ F_{1}$ and $F_{2}$ are the 
adjacent sampled triangular  faces. 
To reduce color difference, we also 
bring in non-printability score (NPS) \cite{thys2019fooling} as 
follows:
\begin{equation}
	\mathcal{L}_{NPS}=\sum_{p_{patch}\in P }\min_{c_{printer}\in C} \left | 
	p_{patch}-c_{printer} \right | 
	\label{13}
\end{equation}
where $c_{printer}$ is a color vector of printable RGB triplets set $C$, 
$p_{patch}$ indicates the color vector of one sampled triangular face, and $P$ 
denotes the set of sampled faces. 

On the whole, we obtain transferable physical adversarial pattern by taking 
separable attention loss, smooth loss and 
non-printability score into consideration simultaneously. 
Therefore, our adversarial pattern is updated by optimizing the 
following formula:
\begin{equation}
	\mathcal{L}=\mathcal{L}_{FAS}+ \mathcal{L}_{BAA}+ 
	\beta \cdot \mathcal{L}_{S,3D}+ \gamma \cdot  \mathcal{L}_{NPS} 
	\label{14}
\end{equation}
where  $\beta$ and $\gamma$ are hyperparameters.  

\section{Experiments and Results}
\label{others}
In this section, we first present experimental settings, and then conduct 
extensive experimental evaluations about black-box transferability to verify 
the effectiveness of our proposed method. Afterwards, a serial of ablation 
studies 
are launched to support further analysis of our method.

\subsection{Experimental Settings}\label{Settings}

\paragraph{Data Collection.}
Both training set and testing set are collected with 
CARLA \cite{dosovitskiy2017carla}, which is an 
open-source 3D simulator developed for autonomous driving research. 
Unlike some works where images are captured at relatively nearby 
locations \cite{wang2021dual,jiang2021fca}, 
we carry out attack on a more complicated dataset
 with longer camera-object 
distances.   
Specifically, we take photos with five distance 
values (R=10, 20, 30, 
40 and 50), three camera pitch values ($pitch$=50, 70 and 90) and eight 
camera yaw values ($yaw$=0, 45, 90, 135, 180, 225, 270 and 315). In other 
words, we get 120 images for a single 
vehicle location (24 images per distance value). To increase diversity of the 
dataset, we randomly select 31 points to place our vehicle  in the virtual 
environment. Finally, we get 3600 images in total, 3000 of which are set as 
training set and 600 as testing set. 
Besides, all the images are obtained with 
a resolution of 1024$\times$1024 in CARLA and then resized to 608$\times$608 
before fed into models.

\paragraph{Detectors Training.}
We first optimize the camouflage 
pattern in white-box detection models, which consist of 
Yolo-V3\footnote{https://github.com/bubbliiiing/yolo3-pytorch} 
\cite{redmon2018yolov3},
Yolo-V5\footnote{https://github.com/ultralytics/yolov5}, Faster 
R-CNN \cite{ren2015faster} and Retinanet \cite{lin2017focal} (Faster R-CNN and 
Retinanet are implemented by
Pytorch). We conduct 
evaluation experiments on six black-box models, which include two classic 
one-stage detectors (one SSD \cite{liu2016ssd} model with fixed input size of 
512 and one Yolo-V5 as the above white-box model), a FPN-based one-stage 
detector (retinanet \cite{lin2017focal}), a couple 
of two-stage detectors (Faster R-CNN \cite{ren2015faster} and Mask 
R-CNN \cite{he2017mask}) and finally one 
transform-based  detector (Deformable DETR \cite{zhu2020deformable}).  
All the evaluation models (except for Yolo-V5) are based on 
MMDetection \cite{chen2019mmdetection}. 
Besides, all the models above are first trained on VisDrone2019 
dataset \cite{9573394} and then fine-tuned on our dataset captured from CARLA. 

\paragraph{Evaluation Metrics.}
In this paper, we take AP as our first metric like previous 
work \cite{wu2020making}. Given that AP is calculated by allowing for various 
of confidence thresholds, our work also takes attack success rate 
(ASR) \cite{jiang2021fca}  as another metric, which involves only one 
confidence 
threshold.

\paragraph{Baseline Methods.}
To evaluate the effectiveness of our method, we compare our transferability 
from white-box detectors to evaluation models with 
several state-of-the-art works, including Dual Attention Suppression Attack 
(DAS) \cite{wang2021dual}, Full-coverage Camouflage Attack (FCA)
\cite{jiang2021fca} and Attention-guided Transfer Attack 
(ATA) \cite{wu2020boosting}.

\paragraph{Implementation Details.}
All the adversarial patterns are optimized under parameters as follows, the 
batch size is 2, the maximum epochs is 5, and the learning rate of SGD 
optimizer is 0.01. Our experiments are conducted on a desktop with an Intel 
core i9 CPU and one Nvidia RTX-3090 GPU. All 
experiments are carried 
out on the Pytorch (with torch 1.8.0 and 
pytorch3D 0.6.0) framework.

\subsection{Comparison of Transferability}\label{Comparison}

In this section, we compare the performance of our method with that of 
baseline methods. To analyze the performance 
objectively, we reproduce baselines by adopting their critical 
loss functions under the same controlled conditions (\emph{e.g.} 
pattern size, 3D smooth loss, \textit{etc.}). 
Particularly, we do not consider distracting human attention like the original 
DAS where a visually-natural texture  was pasted on the pattern. In 
addition, 
our reproduced ATA permits attacker to yield localized and  
amplitude-unconstrained 
perturbations. 
Figure \ref{tranferability} displays typical comparisons about  detection 
results and attention distraction, which manifest the superiority of our 
method.
Moreover, Table \ref{Transfer_yolov3} illustrates the transferability of 
different 
attack methods from white-box models (in the first column) to black-box 
detectors (in the first row). 
As for criteria, the 
lower the AP, the better the transferability, and it is just contrary for the 
ASR. 
Meanwhile, the results of clean images are also displayed in the 
table (denoted as Raw in the second line).
  
\begin{table*}
	\renewcommand\arraystretch{1.2}  
	\caption{{Transferability of different attack methods from white-box 
			models (the first column) to typical black-box models (the 
			first 
			row). 
			The higher (lower) the ASR (AP), 
			the better the transferability.  }}
	\label{Transfer_yolov3}
	\centering

	\resizebox{\textwidth}{32mm}{  %
		\begin{tabular}{c|c|cc|cc|cc|cc|cc|cc}
			\toprule%
			\multirow{2}{*}{}&
			\multirow{2}{*}{\textbf{Method}} &
			\multicolumn{2}{c|}{\textbf{\underline{\, Faster R-CNN \, }}}&
			\multicolumn{2}{c|}{\textbf{\underline{\,\,\, Mask R-CNN \,\,\,}}}&
			\multicolumn{2}{c|}{\textbf{\underline{ \qquad  Yolo-V5  \qquad 
			}}}&
			\multicolumn{2}{c|}{\textbf{\underline{ \qquad \; SSD \; \qquad 
			}}}& 
			\multicolumn{2}{c|}{\textbf{\underline{\quad \; Retinanet \quad 
			\;}}}&
			\multicolumn{2}{c}{\textbf{\underline{\, Deformable Detr\,}}} \\
			&  \multicolumn{1}{c|}{}&AP(\%) & ASR(\%) &AP(\%) & ASR(\%) 
			&AP(\%) & 
			ASR(\%)&AP(\%) & ASR(\%)&AP(\%) & ASR(\%)&AP(\%) & ASR(\%) \\
			\midrule
			\textbf{Raw}& --- &89.1 & 0.0 & 92.5 & 0.0 & 85.1 & 0.0 & 84.3 & 
			0.0 & 
			90.1 & 0.0 & 90.3 & 0.0 \\
			\midrule
			\multirow{5}{*}{\textbf{Yolo-V3}}
			
			& \textbf{FCA} &62.1 & 35.5 & 66.5 & 34.5 & 76.3 & 28.8 & 31.5 & 
			76.7 & 
			75.6 & 26.3 & \textbf{63.7} & \textbf{71.7}\\
			&\textbf{DAS} &64.4 & 36.0 & 66.0 & 34.7 & 78.1 & 20.8 & 34.8 & 
			73.4 & 
			75.2 & 29.3 & 68.2 & 66.8 \\
			&\textbf{ATA} & 49.5 & 53.7 & 55.8 & 43.0 & 76.8 & 9.7 & 41.3 & 
			62.8 & 
			57.7 & 53.0 & 78.4 & 43.3 \\
			&\textbf{Ours} & \textbf{20.6} & \textbf{83.3} & \textbf{33.8} & 
			\textbf{68.3} & \textbf{70.7} & \textbf{31.0} & \textbf{28.2} 
			&\textbf{ 
				80.5} & 
			\textbf{44.9} & \textbf{72.2} & 68.0 & 68.5 \\
			\midrule
			
			\multirow{4}{*}{\textbf{Yolo-V5}}
			
			& \textbf{FCA} & 38.1 & 68.2 & 44.2 & 59.8 & 51.1 & 63.0 & 31.5 & 
			75.3 & 
			\textbf{63.1} & \textbf{61.0} & 63.6 & 70.8 \\
			&\textbf{DAS} & 37.8 & 66.0 & 44.3 & 57.5 & 58.4 & 54.0 & 37.0 & 
			66.8 & 
			65.7 & 57.7 & 65.2 & 66.7 \\
			&\textbf{ATA} & 77.0 & 12.2 & 81.9 & 6.2 & 82.3 & 0.8 & 62.4 & 
			40.0 & 81.5 
			& 4.7 & 82.8 & 3.8 \\
			&\textbf{Ours} & \textbf{31.7} & \textbf{77.3} & \textbf{32.9} & 
			\textbf{72.0} & \textbf{44.2} & \textbf{78.2} & \textbf{17.7} & 
			\textbf{86.4} & 69.0 & 53.3 & \textbf{59.7} & \textbf{83.3} \\
			\midrule
			
			\multirow{5}{*}{\textbf{Faster R-CNN}}
			& \textbf{FCA} & 16.6 & 88.5 & 32.0 & 69.8 & \textbf{75.1} & 
			\textbf{12.7} 
			& 34.3 & 71.0 & 
			\textbf{32.7} & \textbf{84.7} & 76.5 & 50.8\\
			&\textbf{DAS} & 20.3 & 84.8 & 35.1 & 65.7 & 75.6 & 10.5 & 38.6 & 
			67.7 & 
			33.9 & 80.5 & 76.8 & 44.8 \\
			& \textbf{ATA} & 44.0 & 58.3 & 55.5 & 41.3 & 77.5 & 2.0 & 46.5 & 
			59.1 & 
			51.2 & 58.7 & 81.8 & 16.8\\
			& \textbf{Ours} & \textbf{12.6} & \textbf{92.7} & \textbf{27.3} & 
			\textbf{77.0} & 77.1 & 12.3 & \textbf{24.9} & \textbf{81.9} & 
			40.9 & 77.0 & \textbf{74.8} & \textbf{56.2}\\
			
			\midrule
			
			\multirow{5}{*}{\textbf{Retinanet}}
			&  \textbf{FCA} & 13.5 & 88.8 & 28.4 & 72.2 & 75.3 & 13.8 & 
			\textbf{29.9} 
			& 74.8 & 
			26.7 & 85.0 & 78.6 & 45.5\\
			& \textbf{DAS} & 19.0 & 87.2 & 37.1 & 65.7 & 75.8 & 11.0 & 31.2 & 
			\textbf{75.7} & 
			38.6 & 79.0 & 79.4 & 44.8 \\
			&  \textbf{ATA} & 27.3 & 77.7 & 37.0 & 64.3 & 78.7 & 7.3 & 39.7 & 
			66.4 & 
			33.4 & 82.0 & 79.3 & 41.7\\
			& \textbf{Ours} & \textbf{15.3} & \textbf{89.8} & \textbf{28.2} & 
			\textbf{75.2} & \textbf{74.2} & \textbf{15.3} & 31.1 & 72.7 & 
			\textbf{24.8} & \textbf{85.8} & \textbf{75.9} & \textbf{52.2} \\
			\bottomrule
		\end{tabular}
	}
\end{table*}

According to the third row of Table \ref{Transfer_yolov3},  our method 
outperforms the rest 
methods when transferring from Yolo-V3 to both two-stage detectors and 
one-stage 
detectors. Specifically, the ASR (AP) of our method is roughly double (half) 
that of FCA and DAS when attacking Faster R-CNN, Mask R-CNN and 
Retinanet. In addition, the ASR (AP) of our method still takes the lead when  
attacking Yolo-V5 and SSD. Although our method does not have the best 
transferability when attacking Deformable Detr, our ASR (68.5\%) is only about 
3\% smaller than that 
of FCA.

Similarly, according to the APs and ASRs in the fourth row, our method remains 
overwhelmingly superior when transferring from Yolo-V5 to other models. 
Especially, our ASRs could be about 10\% higher than that of baseline methods 
for all the models except for Retinanet. 
The fifth row represents the transferability of adversarial patches generated 
with white-box detector of Faster R-CNN, and the results reveals that our 
method achieves the best performance when attacking both of the two-stage 
detectors, SSD and Deformable Detr. Meanwhile, FCA gets two first places (when 
transferring to Yolo-V5 and Retinanet) despite of its simplicity.
In addition, the comparisons of the last row once again highlight the 
superiority of our methods. 
 
\subsection{Effect of Averaging on Attention Map}

\begin{figure}
	\centering
	\includegraphics[scale=0.28]{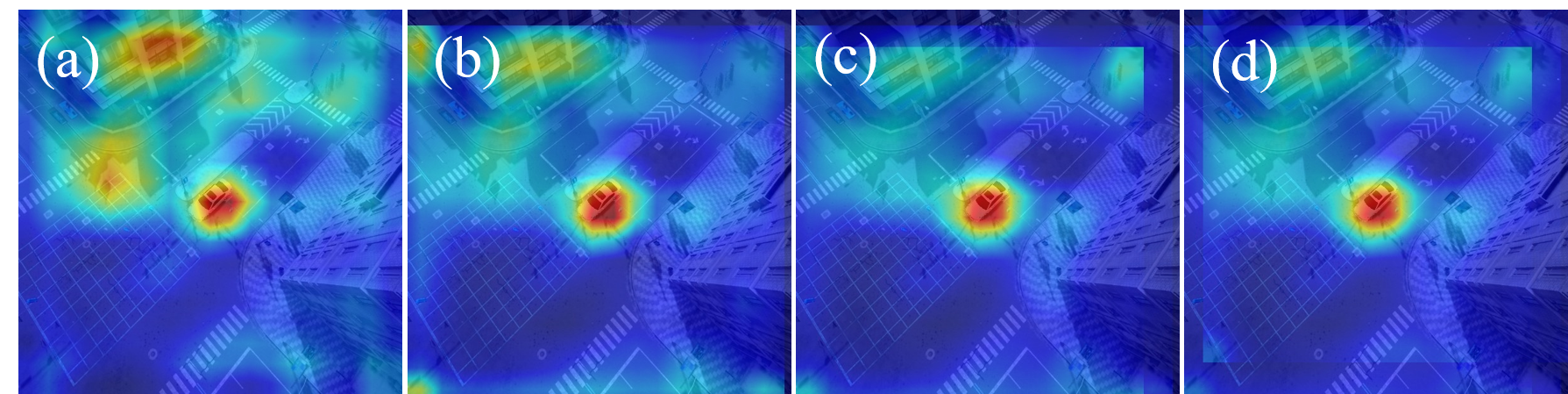}	
	\caption{Effect of compound transformations on attention map.
		(a)$\sim$(d) 
		are the averaged attention maps with  0$\sim$3 compound 
		transformations 
		respectively. 
	} 
	\label{numOfTrans}
\end{figure}

As the inherent noise would be model-specific and hinder 
adversarial examples from transferring to black-box models. To suppress the 
noise, we introduce averaged attention maps.
Figure \ref{numOfTrans}(b)$\sim$(d) display the averaged attention maps with 
1$\sim$3 compound transformations   
respectively, where "heated" regions of the background are alleviated 
noticeably
compared with that of Figure \ref{numOfTrans}(a).

\begin{figure}
	\centering
	\includegraphics[scale=0.24]{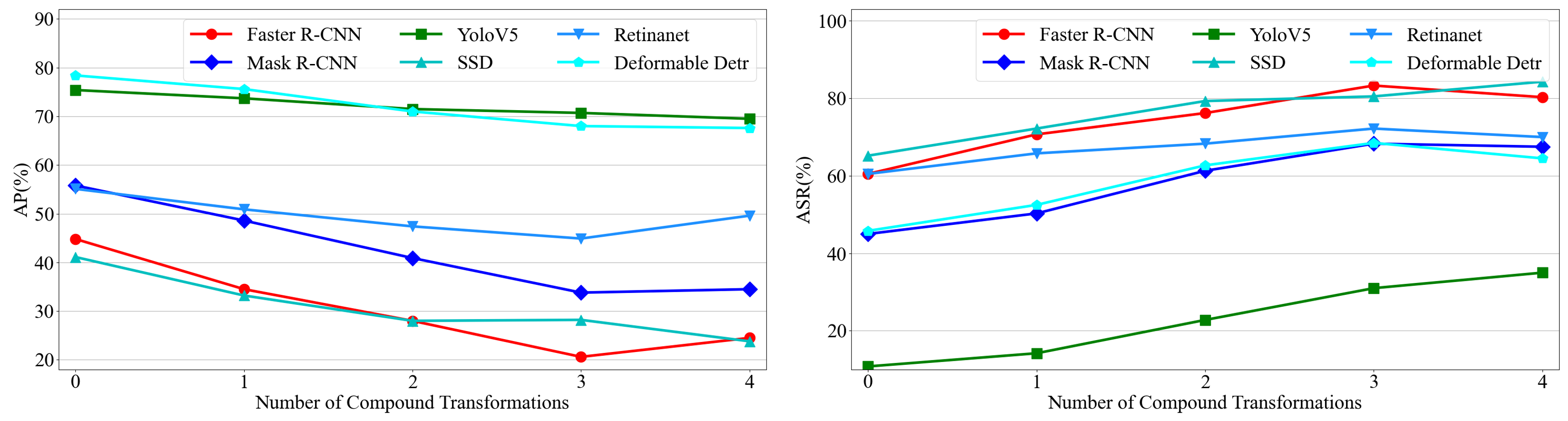}
	\caption{Effect of image transformations on attack transferability. The 
	left (right) graph denotes 
		the correlation between APs (ASRs) and number of compound 
		transformations. 
	}\label{NoOfTrans}
\end{figure}

To verify the effectiveness of averaged attention on transferability, 
 we compute all the APs and ASRs of adversarial patterns 
manufactured with different transformations, covering from zero 
transformation (\textit{i.e.} original images) to 
four compound transformations, each of which includes horizontal flip, 
random scaling and random translation. The final result is demonstrated in 
Figure \ref{NoOfTrans}, where the left graph denotes  correlation between APs  
and number of compound transformations for different black-box models, and the 
right graph indicates the 
variation of ASRs under identical conditions. According to the left graph, 
 APs show a downward trend as we 
increase the number of compound transformation from zero to three (except for 
a 
slight difference for SSD under 3 transformations), and a conspicuous 
turning point could be seen for Retinanet and Faster R-CNN if we continue to 
increase transformation number. On the contrary, the ASR curves in the right 
graph 
witness an opposite variation tendency, \emph{i.e.} 
ASRs go up along with the increase of transformations and the highest point 
can be achieved at 3 compound transformations for most models.  

The experiment above hints that attack transferability could be 
enhanced strikingly (ASRs could be enhanced by about 20\% and APs reduced by 
more than 10\%) by smoothing  attention maps. Actually, a 
well-trained detector could always pay its attention on 
objects no matter how we adjust the input images with basic 
transformations, but the location of noise might be diverse for different 
transformations. Therefore, our method has limited effect on the 
foreground attention but could suppress background noise, which 
finally brings about the enhancement of attack 
transferability.

\subsection{Ablation Studies}\label{Ablation}

In this subsection, we conduct ablation studies to figure out how
hyper-parameters effect transferability, as well as the effect of different 
attention loss items. 

\paragraph{The effect of hyper-parameters on transferability.}

According to equation (\ref{11}), $\alpha_{1}$ is used for controlling the  
global average value of separable attention, and  
$\alpha_{2}$ is devised to adjust the local attention with top $k$ pixels. We 
evaluate the effect of those hyper-parameters with variable-controlling 
method and the results are displayed in Table \ref{Hyper-parameter}.
The second row denotes the effect of $\alpha_{1}$ (=1.0, 2.5, 5.0, 10) on 
transferability with  
$\alpha_{2}=0$, from where we can see that the best performance could be 
achieved when $\alpha_{1}=5$ (except for transferring to Yolo-V5). The third 
row displays 
the impact of $\alpha_{2}$ on  transferability with  $\alpha_{1}=0$ and 
$k=100$. By comparing the results of the second row with that of the third 
row, it is not hard to find that the existence of  $\alpha_{2}$ could enhance 
the attack performance of Yolo-V5, SSD and Deformable Detr. Besides, the last 
row ($\alpha_{1}=0$, $\alpha_{2}=1$) gives a guidance of choosing an 
appropriate $k$ ($=100$). 

\begin{table*}
	\renewcommand\arraystretch{1.1}  %
	\caption{{The comparison of transferability with different hyper  
	parameters. The higher the ASR (or the lower the AP), the better 
			the transferability.  }}
	\label{Hyper-parameter}
	\centering
	
	\resizebox{\textwidth}{25mm}{  %
		\begin{tabular}{c|cc|cc|cc|cc|cc|cc}
			\toprule%
			\multirow{2}{*}{\textbf{Hyper-parameter}} &
			\multicolumn{2}{c|}{\textbf{\underline{\, Faster R-CNN \, }}}&
			\multicolumn{2}{c|}{\textbf{\underline{\,\,\, Mask R-CNN \,\,\,}}}&
			\multicolumn{2}{c|}{\textbf{\underline{ \qquad  Yolo-V5  \qquad 
			}}}&
			\multicolumn{2}{c|}{\textbf{\underline{ \qquad \; SSD \; \qquad 
			}}}& 
			\multicolumn{2}{c|}{\textbf{\underline{\quad \; Retinanet \quad 
						\;}}}&
			\multicolumn{2}{c}{\textbf{\underline{\, Deformable Detr\,}}} \\

			&AP(\%) & ASR(\%) &AP(\%) & ASR(\%) 
			&AP(\%) & 
			ASR(\%)&AP(\%) & ASR(\%)&AP(\%) & ASR(\%)&AP(\%) & ASR(\%) \\
			\midrule
	
			$\alpha_{1}=1.0$ & 
			45.5 & 61.3 & 50.4 & 54.2 & 73.4 & 31.2 & 29.8 & 74.4 & 65.1 & 
			48.0 & \textbf{65.0} & \textbf{71.5} \\
			$\alpha_{1}=2.5$ & 
			27.7 & 77.7 & 37.4 & 64.8 & \textbf{69.8} & \textbf{35.2} & 33.0 & 
			72.7 & 49.8 & 
			69.7 & 65.1 & 66.2\\
			$\alpha_{1}=5.0$ & 
			\textbf{23.7} & \textbf{81.8} & \textbf{35.1} &\textbf{ 65.7} & 
			70.8 & 30.5 & \textbf{27.5} & \textbf{79.1} & \textbf{46.5} & 
			\textbf{73.8} & 67.3 & \textbf{71.5} \\
			$\alpha_{1}=10.0$ & 
			29.2 & 75.7 & 43.6 & 59.7 & 72.9 & 17.5 & 32.1 & 73.9 & 50.5 & 
			69.3 & 74.5 & 60.8 \\
			
			\midrule			
			$\alpha_{2}=0.01$ & 
			\textbf{63.8} & \textbf{35.2} & 66.1 & 33.3 & 79.6 & 16.7 & 27.4 & 
			77.6 & \textbf{74.2} & \textbf{32.0} & 65.3 & 74.8\\
			$\alpha_{2}=0.1$ & 
			65.0 & 31.3 & \textbf{65.4} & \textbf{36.5} & 77.1 & 28.5 & 24.0 & 
			82.4 & 75.1 & 28.2 & 64.4 & \textbf{76.8} \\
			$\alpha_{2}=1.0$ & 
			69.0 & 24.8 & 71.4 & 27.5 & 75.6 & \textbf{37.7} & \textbf{13.6} & 
			93.6 & 77.7 & 	18.2 & \textbf{64.3} & 76.5 \\
			$\alpha_{2}=10.0$ & 
		74.4 & 13.5 & 77.7 & 13.2 &\textbf{ 74.7} & 30.7 & 14.6 & 
		\textbf{94.1} & 81.0 & 7.5 & 71.6 & 68.8 \\
			
						\midrule			
			$k=50$ & 
			73.0 & 20.0 & 75.6 & 19.7 & 75.3 & 31.8 & \textbf{10.7} & 
			\textbf{95.5} & 79.2 & 	15.0 & 68.4 & 73.2 \\
			$k=100$ & 
			\textbf{69.0} & \textbf{24.8} & \textbf{71.4} & \textbf{27.5} & 
			75.6 & \textbf{37.7} & 13.6 & 	93.6 & \textbf{77.7} & 
			\textbf{18.2} & \textbf{64.3} & \textbf{76.5} \\
			$k=150$ & 
			72.1 & 20.8 & 73.8 & 23.5 & \textbf{75.2} & 33.7 & 12.1 & 93.6 & 
			78.8 & 14.5 & 68.8 & 73.5 \\
			$k=200$ & 
			72.0 & 20.0 & 72.9 & 23.2 & 75.8 & 37.2 & 16.4 & 92.9 & 78.4 & 
			15.2 & 68.9 & 75.2 \\
			\bottomrule
		\end{tabular}
	}
\end{table*}

\paragraph{The effect of different attention loss items.}
\begin{table*}
	\renewcommand\arraystretch{1.1}  %
	\caption{{The comparison of transferability under different attention loss 
			items.  }}
	\label{attention_loss}
	\centering
	
	\resizebox{\textwidth}{10mm}{  
		\begin{tabular}{c|cc|cc|cc|cc|cc|cc}
			\toprule%
			\multirow{2}{*}{\textbf{Attention Loss}} &
			\multicolumn{2}{c|}{\textbf{\underline{\, Faster R-CNN \, }}}&
			\multicolumn{2}{c|}{\textbf{\underline{\,\,\, Mask R-CNN \,\,\,}}}&
			\multicolumn{2}{c|}{\textbf{\underline{ \qquad  Yolo-V5  \qquad 
			}}}&
			\multicolumn{2}{c|}{\textbf{\underline{ \qquad \; SSD \; \qquad 
			}}}& 
			\multicolumn{2}{c|}{\textbf{\underline{\quad \; Retinanet \quad 
						\;}}}&
			\multicolumn{2}{c}{\textbf{\underline{\, Deformable Detr\,}}} \\%
			&AP(\%) & ASR(\%) &AP(\%) & ASR(\%) 
			&AP(\%) & 
			ASR(\%)&AP(\%) & ASR(\%)&AP(\%) & ASR(\%)&AP(\%) & ASR(\%) \\
			\midrule
			
			
			\textbf{$\mathcal{L}_{F A S}$} & 
			28.2 & 76.2 & 41.0 & 61.7 & \textbf{69.1} & 29.0 & 26.6 & 78.4 & 
			50.9 & 	66.8 & 68.9 & 64.8 \\
			\textbf{$\mathcal{L}_{B A A}$} & 
			70.7 & 24.2 & 73.7 & 22.8 & 77.1 & 26.7 & \textbf{20.8} & 
			\textbf{88.7} & 77.3 & 20.0 & 68.1 & \textbf{71.2} \\
			\textbf{$\mathcal{L}_{F A S}+\mathcal{L}_{B A A}$} & 
			\textbf{20.6} & \textbf{83.3} & \textbf{33.8} & \textbf{68.3} & 
			70.7 & \textbf{31.0} & 28.2 & 80.5 & \textbf{44.9} & 
			\textbf{72.2} & \textbf{68.0} & 68.5 \\
			
			\bottomrule
		\end{tabular}
	}
\end{table*}

To verify whether a single foreground or background attention 
loss could make a difference, we carry out ablation studies on the effect of 
different attention loss items. Concretely, adversarial patterns are optimized 
under attention loss of $\mathcal{L}_{F A S}$, $\mathcal{L}_{B A A}$ and 
$\mathcal{L}_{F A S}+\mathcal{L}_{B A A}$ respectively
(with $\alpha_{1}=5$, $\alpha_{2}=1$ and $k=100$).
According to Table 
\ref{attention_loss}, the combination of $\mathcal{L}_{F A S}$ and 
$\mathcal{L}_{B A A}$ could contribute to the best performance (both AP 
and ASR) for Faster R-CNN, Mask R-CNN and Retinanet. Meanwhile, both ASR of 
Yolo-V5 (31.0\%) and AP of Deformable Detr (68.0\%) take up the first place 
according to the last row of Table \ref{attention_loss}. Besides, the results 
in the third row show that the single 
$\mathcal{L}_{B A A}$ would
bring about a weaker attack performance among black-box models, with the 
exception of SSD and Deformable Detr. Although  $\mathcal{L}_{F A S}$ 
does not lead to a striking result, its gap from the integrated loss 
(\textit{i.e.} the difference between  the second row and the last row) is 
much smaller than that of $\mathcal{L}_{B A A}$ for the majority detectors, 
which means that foreground attention loss plays a more important role than 
the background attention loss.

\section{Conclusions}\label{conclusions}

In this work, we propose a framework of generating 3D physical adversarial 
camouflage against object detection by attacking multi-scale 
attention maps. As the noise of those maps would be model-specific 
and hinder the manufactured adversarial pattern from transferring to black-box 
detectors, our method first takes serial compound transformations on input 
images, and then aligns the yielded attention maps with inverse 
transformations. Afterwards, we suppress the noise by averaging the aligned 
maps directly. 
 In order to 
break through the limitations of existing attention-based attack methods, 
we propose a direct and efficient optimization strategy based on separable 
attention, and develop a 
novel attention loss function which suppresses the foreground attention and 
amplifies that of the background simultaneously. Moreover, 
extensive comparison experiments are carried out to get transferability of 
black-box 
detectors, and the results show that our method has a notable advantage over 
other advanced attack methods. Last but not least, there still remain some 
models that are hard to be attacked (\textit{e.g.} Yolo-V5),  the solution of 
which will be explored in our future work.

\medskip

{
\small

\bibliographystyle{unsrt}
\bibliography{references} 

\begin{thebibliography}{10}

\bibitem{redmon2018yolov3}
Joseph Redmon and Ali Farhadi.
\newblock Yolov3: An incremental improvement.
\newblock {\em arXiv preprint arXiv:1804.02767}, 2018.

\bibitem{zhu2020deformable}
Xizhou Zhu, Weijie Su, Lewei Lu, Bin Li, Xiaogang Wang, and Jifeng Dai.
\newblock Deformable detr: Deformable transformers for end-to-end object
  detection.
\newblock {\em arXiv preprint arXiv:2010.04159}, 2020.

\bibitem{ren2015faster}
Shaoqing Ren, Kaiming He, Ross Girshick, and Jian Sun.
\newblock Faster r-cnn: Towards real-time object detection with region proposal
  networks.
\newblock {\em Advances in neural information processing systems}, 28, 2015.

\bibitem{gong2019cnn}
Zhiqiang Gong, Ping Zhong, Yang Yu, Weidong Hu, and Shutao Li.
\newblock A cnn with multiscale convolution and diversified metric for
  hyperspectral image classification.
\newblock {\em IEEE Transactions on Geoscience and Remote Sensing},
  57(6):3599--3618, 2019.

\bibitem{gong2020statistical}
Zhiqiang Gong, Ping Zhong, and Weidong Hu.
\newblock Statistical loss and analysis for deep learning in hyperspectral
  image classification.
\newblock {\em IEEE Transactions on Neural Networks and Learning Systems},
  32(1):322--333, 2020.

\bibitem{he2017mask}
Kaiming He, Georgia Gkioxari, Piotr Doll{\'a}r, and Ross Girshick.
\newblock Mask r-cnn.
\newblock In {\em Proceedings of the IEEE international conference on computer
  vision}, pages 2961--2969, 2017.

\bibitem{goodfellow2014explaining}
Ian~J Goodfellow, Jonathon Shlens, and Christian Szegedy.
\newblock Explaining and harnessing adversarial examples.
\newblock {\em arXiv preprint arXiv:1412.6572}, 2014.

\bibitem{wu2020boosting}
Weibin Wu, Yuxin Su, Xixian Chen, Shenglin Zhao, Irwin King, Michael~R Lyu, and
  Yu-Wing Tai.
\newblock Boosting the transferability of adversarial samples via attention.
\newblock In {\em Proceedings of the IEEE/CVF Conference on Computer Vision and
  Pattern Recognition}, pages 1161--1170, 2020.

\bibitem{wang2021feature}
Zhibo Wang, Hengchang Guo, Zhifei Zhang, Wenxin Liu, Zhan Qin, and Kui Ren.
\newblock Feature importance-aware transferable adversarial attacks.
\newblock In {\em Proceedings of the IEEE/CVF International Conference on
  Computer Vision}, pages 7639--7648, 2021.

\bibitem{huang2020universal}
Lifeng Huang, Chengying Gao, Yuyin Zhou, Cihang Xie, Alan~L Yuille, Changqing
  Zou, and Ning Liu.
\newblock Universal physical camouflage attacks on object detectors.
\newblock In {\em Proceedings of the IEEE/CVF Conference on Computer Vision and
  Pattern Recognition}, pages 720--729, 2020.

\bibitem{su2019one}
Jiawei Su, Danilo~Vasconcellos Vargas, and Kouichi Sakurai.
\newblock One pixel attack for fooling deep neural networks.
\newblock {\em IEEE Transactions on Evolutionary Computation}, 23(5):828--841,
  2019.

\bibitem{yang2020learning}
Jiancheng Yang, Yangzhou Jiang, Xiaoyang Huang, Bingbing Ni, and Chenglong
  Zhao.
\newblock Learning black-box attackers with transferable priors and query
  feedback.
\newblock {\em Advances in Neural Information Processing Systems},
  33:12288--12299, 2020.

\bibitem{wang2021dual}
Jiakai Wang, Aishan Liu, Zixin Yin, Shunchang Liu, Shiyu Tang, and Xianglong
  Liu.
\newblock Dual attention suppression attack: Generate adversarial camouflage in
  physical world.
\newblock In {\em Proceedings of the IEEE/CVF Conference on Computer Vision and
  Pattern Recognition}, pages 8565--8574, 2021.

\bibitem{jiang2021fca}
Tingsong Jiang, Jialiang Sun, Weien Zhou, Xiaoya Zhang, Zhiqiang Gong, Wen Yao,
  Xiaoqian Chen, et~al.
\newblock Fca: Learning a 3d full-coverage vehicle camouflage for multi-view
  physical adversarial attack.
\newblock {\em arXiv preprint arXiv:2109.07193}, 2021.

\bibitem{chen2020universal}
Sizhe Chen, Zhengbao He, Chengjin Sun, Jie Yang, and Xiaolin Huang.
\newblock Universal adversarial attack on attention and the resulting dataset
  damagenet.
\newblock {\em IEEE Transactions on Pattern Analysis and Machine Intelligence},
  2020.

\bibitem{eykholt2018robust}
Kevin Eykholt, Ivan Evtimov, Earlence Fernandes, Bo~Li, Amir Rahmati, Chaowei
  Xiao, Atul Prakash, Tadayoshi Kohno, and Dawn Song.
\newblock Robust physical-world attacks on deep learning visual classification.
\newblock In {\em Proceedings of the IEEE conference on computer vision and
  pattern recognition}, pages 1625--1634, 2018.

\bibitem{thys2019fooling}
Simen Thys, Wiebe Van~Ranst, and Toon Goedem{\'e}.
\newblock Fooling automated surveillance cameras: adversarial patches to attack
  person detection.
\newblock In {\em Proceedings of the IEEE/CVF conference on computer vision and
  pattern recognition workshops}, pages 0--0, 2019.

\bibitem{maesumi2021learning}
Arman Maesumi, Mingkang Zhu, Yi~Wang, Tianlong Chen, Zhangyang Wang, and
  Chandrajit Bajaj.
\newblock Learning transferable 3d adversarial cloaks for deep trained
  detectors.
\newblock {\em arXiv preprint arXiv:2104.11101}, 2021.

\bibitem{sharif2016accessorize}
Mahmood Sharif, Sruti Bhagavatula, Lujo Bauer, and Michael~K Reiter.
\newblock Accessorize to a crime: Real and stealthy attacks on state-of-the-art
  face recognition.
\newblock In {\em Proceedings of the 2016 acm sigsac conference on computer and
  communications security}, pages 1528--1540, 2016.

\bibitem{wu2020making}
Zuxuan Wu, Ser-Nam Lim, Larry~S Davis, and Tom Goldstein.
\newblock Making an invisibility cloak: Real world adversarial attacks on
  object detectors.
\newblock In {\em European Conference on Computer Vision}, pages 1--17.
  Springer, 2020.

\bibitem{wang2021towards}
Yajie Wang, Haoran Lv, Xiaohui Kuang, Gang Zhao, Yu-an Tan, Quanxin Zhang, and
  Jingjing Hu.
\newblock Towards a physical-world adversarial patch for blinding object
  detection models.
\newblock {\em Information Sciences}, 556:459--471, 2021.

\bibitem{du2022physical}
Andrew Du, Bo~Chen, Tat-Jun Chin, Yee~Wei Law, Michele Sasdelli, Ramesh
  Rajasegaran, and Dillon Campbell.
\newblock Physical adversarial attacks on an aerial imagery object detector.
\newblock In {\em Proceedings of the IEEE/CVF Winter Conference on Applications
  of Computer Vision}, pages 1796--1806, 2022.

\bibitem{athalye2018synthesizing}
Anish Athalye, Logan Engstrom, Andrew Ilyas, and Kevin Kwok.
\newblock Synthesizing robust adversarial examples.
\newblock In {\em International conference on machine learning}, pages
  284--293. PMLR, 2018.

\bibitem{tarchoun2021adversarial}
Bilel Tarchoun, Ihsen Alouani, Anouar~Ben Khalifa, and Mohamed~Ali Mahjoub.
\newblock Adversarial attacks in a multi-view setting: An empirical study of
  the adversarial patches inter-view transferability.
\newblock In {\em 2021 International Conference on Cyberworlds (CW)}, pages
  299--302. IEEE, 2021.

\bibitem{xiang2021local}
Tao Xiang, Hangcheng Liu, Shangwei Guo, Tianwei Zhang, and Xiaofeng Liao.
\newblock Local black-box adversarial attacks: A query efficient approach.
\newblock {\em arXiv preprint arXiv:2101.01032}, 2021.

\bibitem{guo2019simple}
Chuan Guo, Jacob Gardner, Yurong You, Andrew~Gordon Wilson, and Kilian
  Weinberger.
\newblock Simple black-box adversarial attacks.
\newblock In {\em International Conference on Machine Learning}, pages
  2484--2493. PMLR, 2019.

\bibitem{zhou2018transferable}
Wen Zhou, Xin Hou, Yongjun Chen, Mengyun Tang, Xiangqi Huang, Xiang Gan, and
  Yong Yang.
\newblock Transferable adversarial perturbations.
\newblock In {\em Proceedings of the European Conference on Computer Vision
  (ECCV)}, pages 452--467, 2018.

\bibitem{inkawhich2019feature}
Nathan Inkawhich, Wei Wen, Hai~Helen Li, and Yiran Chen.
\newblock Feature space perturbations yield more transferable adversarial
  examples.
\newblock In {\em Proceedings of the IEEE/CVF Conference on Computer Vision and
  Pattern Recognition}, pages 7066--7074, 2019.

\bibitem{huang2019enhancing}
Qian Huang, Isay Katsman, Horace He, Zeqi Gu, Serge Belongie, and Ser-Nam Lim.
\newblock Enhancing adversarial example transferability with an intermediate
  level attack.
\newblock In {\em Proceedings of the IEEE/CVF international conference on
  computer vision}, pages 4733--4742, 2019.

\bibitem{inkawhich2020perturbing}
Nathan Inkawhich, Kevin Liang, Binghui Wang, Matthew Inkawhich, Lawrence Carin,
  and Yiran Chen.
\newblock Perturbing across the feature hierarchy to improve standard and
  strict blackbox attack transferability.
\newblock {\em Advances in Neural Information Processing Systems},
  33:20791--20801, 2020.

\bibitem{chen2022relevance}
Sizhe Chen, Fan He, Xiaolin Huang, and Kun Zhang.
\newblock Relevance attack on detectors.
\newblock {\em Pattern Recognition}, 124:108491, 2022.

\bibitem{ravi2020accelerating}
Nikhila Ravi, Jeremy Reizenstein, David Novotny, Taylor Gordon, Wan-Yen Lo,
  Justin Johnson, and Georgia Gkioxari.
\newblock Accelerating 3d deep learning with pytorch3d.
\newblock {\em arXiv preprint arXiv:2007.08501}, 2020.

\bibitem{dosovitskiy2017carla}
Alexey Dosovitskiy, German Ros, Felipe Codevilla, Antonio Lopez, and Vladlen
  Koltun.
\newblock Carla: An open urban driving simulator.
\newblock In {\em Conference on robot learning}, pages 1--16. PMLR, 2017.

\bibitem{selvaraju2017grad}
Ramprasaath~R Selvaraju, Michael Cogswell, Abhishek Das, Ramakrishna Vedantam,
  Devi Parikh, and Dhruv Batra.
\newblock Grad-cam: Visual explanations from deep networks via gradient-based
  localization.
\newblock In {\em Proceedings of the IEEE international conference on computer
  vision}, pages 618--626, 2017.

\bibitem{lin2017focal}
Tsung-Yi Lin, Priya Goyal, Ross Girshick, Kaiming He, and Piotr Doll{\'a}r.
\newblock Focal loss for dense object detection.
\newblock In {\em Proceedings of the IEEE international conference on computer
  vision}, pages 2980--2988, 2017.

\bibitem{liu2016ssd}
Wei Liu, Dragomir Anguelov, Dumitru Erhan, Christian Szegedy, Scott Reed,
  Cheng-Yang Fu, and Alexander~C Berg.
\newblock Ssd: Single shot multibox detector.
\newblock In {\em European conference on computer vision}, pages 21--37.
  Springer, 2016.

\bibitem{chen2019mmdetection}
Kai Chen, Jiaqi Wang, Jiangmiao Pang, Yuhang Cao, Yu~Xiong, Xiaoxiao Li,
  Shuyang Sun, Wansen Feng, Ziwei Liu, Jiarui Xu, et~al.
\newblock Mmdetection: Open mmlab detection toolbox and benchmark.
\newblock {\em arXiv preprint arXiv:1906.07155}, 2019.

\bibitem{9573394}
Pengfei Zhu, Longyin Wen, Dawei Du, Xiao Bian, Heng Fan, Qinghua Hu, and Haibin
  Ling.
\newblock Detection and tracking meet drones challenge.
\newblock {\em IEEE Transactions on Pattern Analysis and Machine Intelligence},
  pages 1--1, 2021.

\end{thebibliography}
}

\appendix

\end{document}